\documentclass[10pt,twocolumn,letterpaper]{article}

\usepackage[title]{appendix}
\usepackage{cvpr}
\usepackage{times}
\usepackage{epsfig}
\usepackage{graphicx}
\usepackage{amsmath}
\usepackage{amssymb}
\usepackage{multirow}

\usepackage[breaklinks=true,bookmarks=false]{hyperref}

\cvprfinalcopy 

\def\cvprPaperID{****} 

\begin{document}

\title{Patch Correspondences for Interpreting Pixel-level CNNs}

\author{Victor Fragoso$^{2,3}$ \quad Chunhui Liu$^1$ \quad Aayush Bansal$^1$ \quad Deva Ramanan$^1$\\
$^{1}$ Carnegie Mellon University \quad $^2$ Microsoft C + AI  \quad $^3$ Western Virginia University \\
}

\maketitle

\begin{abstract}
We present compositional nearest neighbors (CompNN), a simple approach to visually interpreting distributed representations learned by a convolutional neural network (CNN) for pixel-level tasks (e.g., image synthesis and segmentation). It does so by reconstructing both a CNN's input and output image by copy-pasting corresponding patches from the training set with similar feature embeddings. To do so efficiently, it makes of a patch-match-based algorithm that exploits the fact that the patch representations learned by a CNN for pixel level tasks vary smoothly. Finally, we show that CompNN can be used to establish semantic correspondences between two images and control properties of the output image by modifying the images contained in the training set. We present qualitative and quantitative experiments for semantic segmentation and image-to-image translation that demonstrate that CompNN is a good tool for interpreting the embeddings learned by pixel-level CNNs.
\end{abstract}
\section{Introduction}
\label{sec:introduction}

Convolutional neural networks (CNNs) have revolutionized computer vision because they are excellent mechanisms to learn good representations for solving visual tasks. Recently, they have produced impressive results for discriminative tasks, such as, image classification and semantic segmentation~\cite{Krizhevsky_NIPS2012, Simonyan15, Szegedy_2015_CVPR, He_2016_CVPR, huang2017densely}, and have produced startlingly impressive results for image generation through generative models~\cite{pix2pix2016, chen2017photographic}. While these results are remarkable, these feed-forward networks are sometimes criticized because it is hard to exactly determine what information is encoded to produce great results. Consequently, it is difficult to explain why networks sometimes fail on particular inputs or how they will behave on never-before-seen data. Given this difficulty, there is a renewed interest in {\em explainable AI}\footnote{\url{http://www.darpa.mil/program/explainable-artificial-intelligence}}, which mainly fosters the development of machine learning systems that are designed to be more interpretable and explanatory.

\begin{figure*}[t]
\centering
 \includegraphics[width=1\textwidth]{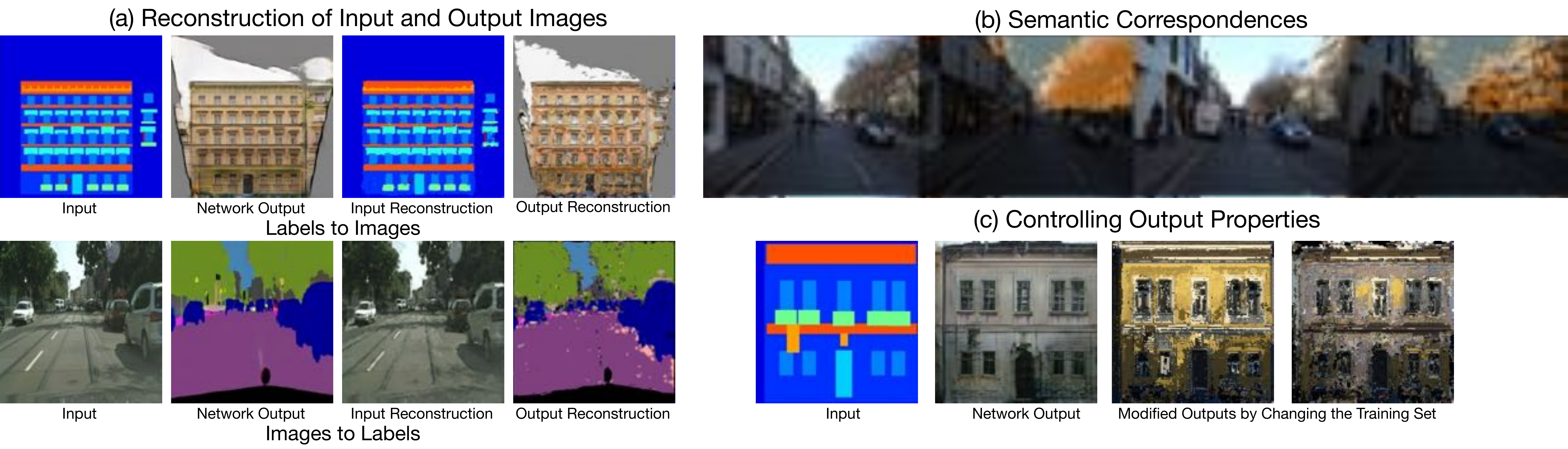}
 \vspace{-4mm}
 \caption{CompNN establishes correspondences between patches from the input image and images of a training set. To compute these correspondences given an input image, CompNN searches for the most similar patches in the training set given the patches that compose the input image. To measure similarity, CompNN uses patch representations extracted from the learned embeddings by CNNs for pixel-level tasks. The correspondences computed by CompNN are useful to {\bf (a)} reconstruct both input and output images by composing the patches and assembling a coherent image, these reconstructions thus can help the user interpret the outputs of a CNN; {\bf (b)} establish semantic correspondences between image pairs; and {\bf (c)} control properties of the output image by including or removing images from the training set, which is useful to understand and possibly correct the implicit bias in a CNN.}
\label{fig:teaser}
\end{figure*}

To better understand the CNNs representations, the computer vision and machine learning communities have developed visualizations that interpret the representations learned by a CNN. These include mechanisms that identify images that maximally excite neurons, reconstruct input images by inverting their representations, identify regions in images that contribute the most to a target response, among others~\cite{krishnan2016tinkering, mahendran2016visualizing, fong2018net2vec, bilal2018convolutional, olah2018building}. Such methods are typically applied to CNNs designed for global tasks such as image classification. Instead, we focus on understanding CNNs trained for detailed pixel-level prediction tasks (e.g., image translation and segmentation~\cite{pix2pix2016, chen2017photographic, wang2017high}), which likely requires different representations that capture local pixel structure. 

Our approach to interpretability is inspired by case-based reasoning or ``explanation-by-example'' ~\cite{lipton2016mythos}. Such an approach dates back to classic AI systems that predict medical diagnoses or legal judgments that were justified through case studies or historical precedent~\cite{aamodt1994case}. For example, radiologists can justify diagnoses of an imaged tumor as `malignant' by reference to a previously-seen example~\cite{caruana1999case}. However, this approach can generate only $N$ explanations given $N$ training exemplars. In theory, one could provide exponentially more explanations through {\em composition}: e.g., this {\em part} of the image looks like this {\em part} of exemplar A, while another {\em part} looks like that {\em part} of exemplar B. We term this ``explanation-by-correspondence,'' since such explanations provide correspondence of patches from a query image to patches from exemplars.

We present CompNN, a simple approach to visually interpret the representations learned by a CNN for pixel-level tasks. Different from existing visual interpretation methods~\cite{krishnan2016tinkering, mahendran2016visualizing, fong2018net2vec, bilal2018convolutional, olah2018building}, CompNN reconstructs a query image by retrieving similar image patches from training exemplars and rearranging them spatially to match the query. To do so, CompNN matches image patches using feature embeddings extracted from different layers of the CNN. Because such a reconstruction provides patch correspondences between the query image and training exemplars, they can also be used to transfer labels from training exemplars. From this perspective, CompNN also provides a reconstruction of the CNN output.
 
While CompNN can operate by exhaustively searching for the most similar patches, it uses a patch-match-like~\cite{barnes2009patchmatch} algorithm that uses patch representations extracted from the learned embeddings by the CNN. This patch-match mechanism allows CompNN to efficiently compute patch correspondences between an input image and images from the training set. CompNN computes these correspondences efficiently because the learned embeddings by the CNN can vary smoothly: an input patch centered at $(x, y)$ with a corresponding patch from a training image centered at $(i, j)$ likely has a neighbor patch centered at $(x + 1, y + 1)$ with a corresponding patch centered at $(i + 1, j + 1)$. This property is crucial for a patch-match-like algorithm since it exploits this property to speed-up the nearest-neighbor search.

In contrast to interpreting the embeddings by retrieving the most similar instance from the training set, composing an image by arranging image patches from a training set enables an exponential range of possible images that can be generated. This is because CompNN has at hand $(NK)^K$ image patches from a training set, where $K$ is the number of patches one can extract from an image and $N$ is the number of training images. Furthermore, patch correspondences are useful because not only they enable the reconstruction of both the input and output images, but also allow a user to understand how a CNN may behave on never-before-seen data, establish semantic correspondences between a pair of images, and   generate an output image with different properties by changing the set of images in the training set. This latter feature is useful to understand and possibly correct the implicit bias in a CNN. As an illustrative example, we can synthesize images with CompNN depicting European or American building facades by restricting the set of images used for computing patch correspondences. Fig.~\ref{fig:teaser} gives an overview of CompNN.


\section{Related Work}
We broadly classify networks for pixel-level tasks or spatial prediction into two categories: (1) discriminative prediction, where the goal is to infer high-level semantic information from RGB values; and (2) image generation, where the intent is to synthesize a new image from a given input ``prior.'' There is a broad literature for each of these tasks, and here we discuss the ones most relevant to ours. We also discuss methods that help users to interpret visual representations.

\noindent\textbf{Discriminative models: } An influential formulation for state-of-the-art spatial prediction tasks is that of fully convolutional networks~\cite{Long15}. These networks have been used for pixel prediction problems such as semantic segmentation~\cite{Long15,Hariharan15,RonnebergerFB15,PixelNet,ChenPK0Y16}, depth and surface-normal estimation~\cite{Bansal16, Eigen15}, and low-level edge detection~\cite{Xie15,PixelNet}. Substantial progress has been made to improve the performance by either employing deeper architectures~\cite{he2015deep}, increasing the capacity of the models~\cite{PixelNet},  utilizing skip connections, or intermediate supervision~\cite{Xie15}. However, we do not precisely know what these models are actually capturing to do pixel-level prediction. In the race for better performance, the interpretability of these models has been typically ignored. In this work, we focus on interpreting the learned embeddings by encoder-decoder architectures for spatial classification~\cite{RonnebergerFB15,BadrinarayananK15}.

\noindent\textbf{Image generation: } Goodfellow et al.~\cite{GoodfellowPMXWOCB14} proposed Generative Adversarial Networks (GANs). These networks consist of a two-player min-max formulation, where a generator $G$ synthesizes an image from random noise $z$, and a discriminator $D$ distiguishes the synthetic images from the real ones. While the original purpose of GANs is to synthesize an image from random noise vectors $z$, this formulation can also be used to synthesize new images from other priors, such as, a low resolution images or label mask by treating $z$ as an explicit input to be conditioned upon. This conditional image synthesis via a generative adversarial formulation has been well utilized by multiple follow-up works to synthesize a new image conditioned on a low-resolution image~\cite{DentonCSF15}, class labels~\cite{RadfordMC15}, and other inputs~\cite{pix2pix2016, ZhuPIE17, Recycle-GAN}. While the quality of synthesis from different inputs has rapidly improved in recent history, interpretation of GANs has been relatively unexplored. In this work, we examine the influential Pix2Pix network~\cite{pix2pix2016} (a conditional GAN), and demonstrate an intuitive non-parametric method for interpreting the learned embeddings generating its impressive results. 

Besides GAN-based methods for generating images, there exist other efforts that synthesize images using deep-features in combination with existing image-synthesis algorithms. These methods use deep-features as intermediate representations of the visual content that another algorithm (e.g., PatchMatch~\cite{barnes2009patchmatch}) can leverage to sinthesize an image. Liao~\etal~\cite{liao17} proposed an approach that generates a pair of images showing visual attribute transfer given an input image pair using deep image analogies. Their approach first identifies structure of one input image and the style of the second input image. They do so by identifying the  structure and style from deep features computed using a pre-trained CNN. Then, they compute bidirectional patch-match-based correspondences using the features extracted from the CNN to finally produce a pair of images. The result is a pair of images that show an exchange of visual attributes from the input image pair. Li and Wand~\cite{li2016combining} propose a method that combines deep features and a Markov random field (MRF) to generate a pair of images showing the visual content and style of an input image pair transferred. Yang~\etal~\cite{yang2017high} proposes a hole-filling method using also deep features.The main goal of these methods is to generate compelling images for visual style transfer or hole filling. On the other hand, the goal of CompNN is to visualize the reconstruction of the input and output image in order to interpret better the embeddings that pixel-level CNNs learn.

\noindent\textbf{Interpretability: }  There is a substantial body of work~\cite{ZeilerF15,Mahendran_2015,ZhouKLOT14,BauZKOT17} on interpreting general convolutional neural networks (CNNs). The earlier work of Zeiler and Fergus~\cite{ZeilerF15} presented an approach to understand and visualize the functioning of intermediate layers of CNN. Zhou et al.~\cite{ZhouKLOT14} demonstrated that object detectors automatically emerge while learning the representation for scene categories.
Recently, Bau et al.~\cite{BauZKOT17} quantify interpretability by measuring scene semantics, such as objects, parts, texture, material etc.. Despite this work, understanding, the space of pixel-level CNNs is not well studied. The recent work of PixelNN~\cite{pixelnn} focuses on high-quality image synthesis by making use of a two-stage matching process that begins by feed-forward CNN processing, and ends with a nonparametric matching of high-frequency detail. In contrast with PixelNN, we focus on interpretability rather than image synthesis. 
Our approach is most similar to those that visualize features by reconstructing an input image through feature inversion~\cite{vondrick2013hoggles,Mahendran_2015}.
 However, rather than training a separate CNN to perform the feature inversion~\cite{Mahendran_2015}, we use simple patch feature matching to produce a reconstruction with correspondences. Crucially, correspondences allow one for additional diagnostics such as output reconstruction through label transfer.

\noindent\textbf{Compositionality: } The design of part-based models~\cite{Crandall2005,FelzenszwalbMR_CVPR_2008}, pictorial structures or spring-like connections~\cite{Fischler1973,Felzenszwalb2005}, star-constellation models~\cite{weber2000towards,Fergus03objectclass}, and the recent works using CNNs share a common theme of \textit{compositionality}. While such earlier work explicitly enforces the idea of composing different parts for object recognition in the algorithmic formulation, there have been suggestions that CNNs also take a compositional approach~\cite{ZeilerF15,KrishnanR16,BauZKOT17}. Our work builds on such formulations, but takes a {\em non-parametric} approach to composition by composing together patches from different training exemplars. From this perspective, our approach is related to the work of Boiman and Irani~\cite{BoimanI06}.


\section{CompNN: Compositional Nearest Neighbors}
\label{sec:compositional_nn}

\begin{figure*}[t]
\centering
\includegraphics[width=\textwidth]{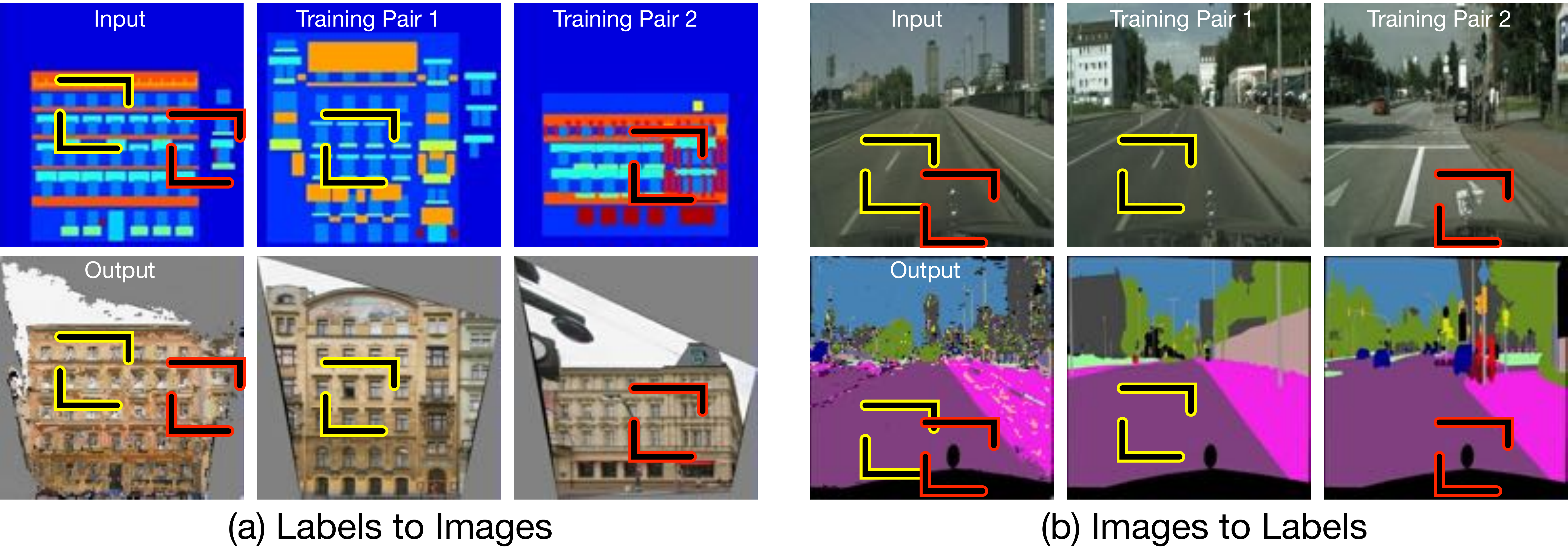}
\vspace{-4mm}
\caption{Overview of the steps in Compositional Nearest Neighbors (CompNN) for {\bf (a)} labels to images and {\bf (b)} images to labels. Given an input patch (double-colored-stroked squares) in the top-left image at the top row, CompNN searches in the training set for the most similar patches  (top row of training pairs 1 \& 2 columns). Then, to reconstruct the output image, CompNN copies the patches from to target images (bottom row of training pairs 1 \& 2 columns) 
and pastes them in the canvas of the output images. CompNN applies the same procedure to reconstruct the input image but copies information from the input training images. Patch correspondences are color coded.}
\label{fig:compnn_overview}
\vspace{-4mm}
\end{figure*}

The general idea of Compositional Nearest Neighbors (CompNN) is to establish patch correspondences between the input image and patches from images in the training set. Given these correspondences, CompNN will sample patches from the training set and assemble them to generate a coherent image. Because our focus is to study encoder-decoder architectures that map an image into another image domain, we assume that the training set contains pairs of images that are aligned pixel-wise. For instance, in the segmentation problem, every pixel in the input image is assigned a class label. Thanks to this setting, CompNN can generate images for resembling the CNN's input and output images.

Establishing patch correspondences requires a representation of the patches and a similarity or distance function. Similar to a Nearest Neighbor (NN) approach, CompNN will search for patches in the training set that are the most similar or proximal in the patch representation space. We use the learned embeddings learned by the network to represent patches and a cosine distance to compare these representations. Fig.~\ref{fig:compnn_overview} illustrates the overall steps that CompNN performs to compute patch correspondences and generating images.

\subsection{Extracting Patch Representations from the CNN Embeddings}
\label{sec:patch_representations}
\begin{figure*}[t]
\centering
\includegraphics[width=\textwidth]{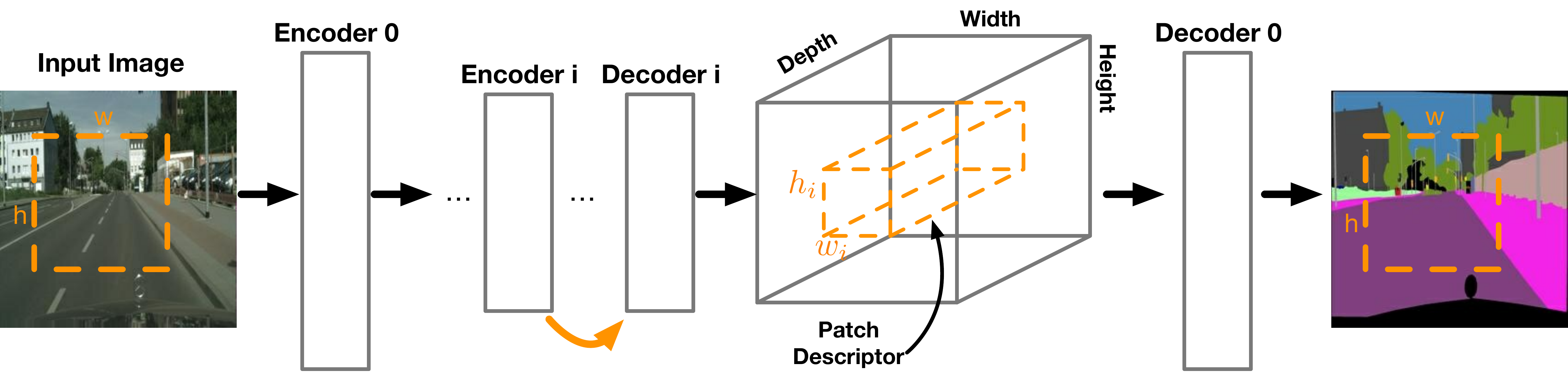}
\vspace{-4mm}
\caption{We extract patch representations from the activation tensors of a particular layer in the network by grabbing ``hyperpatches'' (dashed orange rectangular prism). The width $w_i$ and height $h_i$ of these hyperpatches at layer $i$ depend on the filter sizes of layer $i$ and layer type. To extract a patch representation from a decoder layer, we identify the entries in the activation tensor that contribute to the pixels of the patch in the output image. For instance, a $2 \times 2$ patch in the output image requires a $2 \times 2 \times d_1$ hyperpatch from the activation tensor that is the input to Decoder 0 layer, assuming that Decoder 0 layer applies transposed-convolution using $2 \times 2$ filters. When the hyperpatch belongs to the encoder, then we identify its corresponding decoder layer (orange arrow) and use the same hyperpatch dimensions.}
\label{fig:patch_descriptors}
\vspace{-4mm}
\end{figure*}

A key ingredient of CompNN is the patch representation. We extract patch representations from the activation tensors of a particular layer in a CNN with an encoder-decoder architecture. We assume that a CNN for pixel-level tasks or spatial predictions learns an embedding for every layer in the network. We consider a layer in this work to involve convolutional layers, batch normalizations, and pooling operations. We represent an input image patch with a sub-tensor or ``hyperpatch'' with dimensions $(h_i, w_i, d_i)$ from the activation tensor of the $i$-th layer. Fig.~\ref{fig:patch_descriptors} illustrates the setting and computation of the patch representation as well as the patch in the input image. If the activation tensor for the layer has dimensions of $(H_i, W_i, D_i)$, then $h_i \leq H_i$, $w_i \leq W_i$, and $d_i = D_i$. 

In practice, we first determine the minimal patch size that is constrained by the first encoding layer. For instance, if the first encoder layer convolves with $2 \times 2$ filters, then the patch size of that layer is $2 \times 2$. Then, we identify the corresponding decoder layer of the first encoder layer and we calculate the hyperpatch dimensions as follows. First, we use the activation tensor which is the input to the identified decoder layer. From this activation tensor, we identify the entries that contribute to the production of the output patch. For instance, the hyperpatch dimension to represent a $2 \times 2$ patch in the output image from the last decoder layer is $2 \times 2 \times d_1$, where $d_1$ is the depth of the input tensor to the last decoder layer that uses transposed convolution with $2 \times 2$ filters. Since many architectures downsample by half in their encoder sections, then the patch sizes double in size until reaching the bottleneck of the architecture. For this context, the hyperpatch dimensions for the remaining decoder layers stay the same. For instance, a $4 \times 4$ patch corresponding to the second encoder layer uses a $2 \times 2 \times d_2$ hyperpatch from the activation tensor which is the input to the second to last decoder layer.

\subsection{Computing Patch Correspondences}
\label{sec:patch_correspondences}

The simplest method to establish patch correspondences is by means of an exhaustive search: given an input patch representation, search for the most similiar or proximal patch representation from the same layer that the input patch representation was extracted. As discussed earlier, we use the cosine distance to measure similarity or proximity. This approach mimics 2D convolution. This is because of two reasons. First, this exhaustive search compares the hyperpatch representing a patch in the input image with all possible hyperpatches from an activation tensor at a particular layer. Second, the cosine distance involves a dot product of normalized hypterpatches, which in turn can be implemented as a convolution operation since it is a linear operation. Although this search is highly parallelizable, its drawback is its $\mathcal{O}(h_i w_i d_i)$ computational cost. 

\begin{figure*}[t]
\centering
\includegraphics[width=\textwidth]{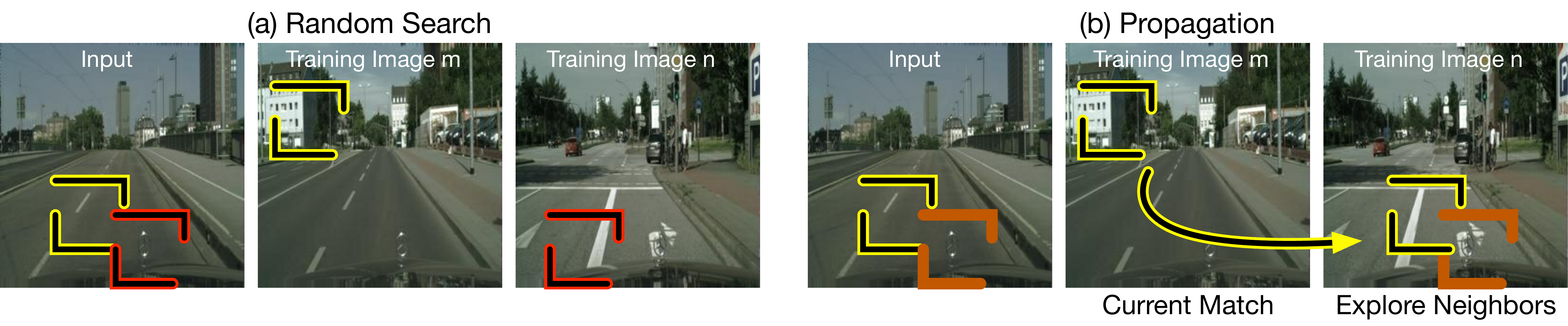}
\vspace{-4mm}
\caption{HyperPatch-Match steps. {\bf (a)} Random search: Given the query patches (top-left squares), this step selects patches and images from the training set at random; this step is used to initialize the algorithm. {\bf (b) } Propagation: Given the query patch in yellow, propagation examines the correspondence of a neighbor patch (orange square). If the neighbor patch (orange square) in image $n$ produces a best similarity value, then propagation updates the correspondence datum for the yellow patch.}
\label{fig:hyperpatch_match}
\vspace{-4mm}
\end{figure*}

To alleviate this computational cost, we approximate the exhaustive search. Inspired by the Patch-Match~\cite{barnes2009patchmatch} algorithm, we developed HyperPatch-Match, a method that approximates the exhaustive search by exploiting the smooth variation that natural scenes possess. This smooth variation ensures with high probability that an image patch from the input image centered at pixel $q_1 = (x, y)$ with a corresponding patch from the training set centered at $t_1 = (i, j)$ has a neighbor patch (e.g., one centered at $q_2 = (x + 1, y + 1)$) with a corresponding patch in the training set that is a neigbhor of the patch centered at $t_1$ (e.g., a neighbor centered at $t_2 = (i + 1, j + 1)$). Unlike Patch-Match that computes patch correspondences between two images in the original input image space (e.g., the RGB space), HyperPatch-Match computes correspondences across a set of several training images with their respective hyperpatches. 

Similar to Patch-Match, HyperPatch-Match has two main steps: random search and propagation. The random search step randomly selects an image from the training set and a patch from the selected image. If the selected patch is most similar to the current similar patch found, then HyperPatch-Match updates the correspondence for a given input patch by storing the id of the training image and the center of the patch. On the other hand, the propagation step uses the smooth variation of natural scenes, which is also maintained in the activation tensors as shown experimentally in Section~\ref{sec:experiments}. Given a query patch from the input image centered at $q = (x, y)$, the propagation step retrieves the current correspondence from a neighbor patch, e.g., $q^\prime = (x + 1, y + 1) \leftrightarrow t=(i, j, l)$, where $(i, j)$ is the center of the patch from the $l$-th training image. Then, the propagation step checks if the cosine distance between $q$ and $t^\prime=(i - i, j - 1, l)$ produce a smaller distance than the current one stored for $q$. If this is the case, the propagation step updates the correspondence datum. HyperPatch-Match initializes the correspondences at random and repeats these steps for several iterations.  While this approach empirically shows faster results than an exhaustive search, HyperPatch-Match potentially still needs to check several images from the training set at random. Note that the random step is the one in charge of exploring new images in the training set, while the propagation step only exploits the smooth variation property. Fig.~\ref{fig:hyperpatch_match} illustrates the steps involved in the proposed HyperPatch-Match method.

To accelerate HyperPatch-Match even more, we used one of the activation tensors from the middle layers as a global image descriptor to select the top $k$ most similar images and utilize them as the training set for a given query input image. To select the top $k$ global nearest neighbors, we compare the global image descriptor of the input image with all the global descriptors of each of the images in the training set, and keep the $k$ most similar images. In this way, HyperPatch-Match reduces the set size of the training images to consider while barely affecting its performance. 

\subsection{CompNN:A Tool for Interpreting Pixel-Level CNN Embeddings}
Similar to other interpretation methods~\cite{ZeilerF15,Mahendran_2015,ZhouKLOT14,BauZKOT17}, CompNN assumes that each layer of the CNN computes an embedding for the input image. Unlike the existing interpretation methods, CompNN focuses on interpreting the embeddings learned by CNNs devised for spatial predictions or pixel-level tasks. CompNN can be interpreted as an inversion method~\cite{mahendran2016visualizing}. This is because CompNN reconstructs the input image given the patch representations extracted from the embeddings learned at every layer. Unlike existing inversion methods that learn functions that recover the input image from a representation, CompNN inverts or reconstructs the input image by exploiting patch correspondences between the input image and images in the training set. While CompNN can reconstruct the input image, it also can generate an output image resembling the output of the CNN. Unlike previous interpretation methods that only focus on understanding representations for image classification, CompNN aims to understand the embeddings that enable the underlying CNN for spatial predictions tasks to synthesize images.

The correspondences computed by CompNN not only are useful to interpret the embeddings of a CNN, but also enable various applications. For instance, the correspondences can be used to establish semantic correspondences between to images, or they can be used to control different properties of the generated output image (e.g., the color and style of a facade) by manipulating the images in the training set. In the next section, we present various results on input reconstructions and output image generation, semantic correspondences given an image pair, and controling properties of the generated output image.
\section{Experiments}
\label{sec:experiments}

In this section, we present a series of qualitative and quantitative experiments to assess the input and output reconstructions, semantic correspondences, and property-controlling of the output reconstruction. For these experiments, we focus on analyzing embeddings learned by the Pix2Pix~
\cite{pix2pix2016} and SegNet~\cite{BadrinarayananK15} networks. The experiments consider image segmentation and image translation as the visual tasks to solve with the aforementioned networks.

\subsection{Input and Output Image Reconstructions}
\label{sec:input_output_recon}
In this section we present qualitative and quantitative experiments that assess the reconstructions for the input and output images of the underlying CNN. For these experiments we used the U-Net-based Pix2Pix~\cite{pix2pix2016} network. Note that the layers of the generator are referenced as Encoder 1-7 following with Decoder 7-1.

To get all the patch representations, we used the publicly available\footnote{Pix2Pix: \url{https://github.com/affinelayer/pix2pix-tensorflow}} pre-trained models for the facades and cityscapes datasets. Because we are interested in interpreting the embeddings that each of the layers of a CNN learn, we extracted all the patch representations for every encoder and decoder layers of the training and validation sets; see Sec.~\ref{sec:patch_representations} for details on how to extract patch representations from the activation tensors at each layer.

Given these patch representations, we used HyperPatch-Match to establish correspondences and reconstruct both input and output images, as described in Section~\ref{sec:patch_correspondences}. We used the top 16 global nearest neighbors to constrain the set used as the training set for every image in the validation set. To select these global nearest neighbors, we used the whole tensor from the Decoder-7 (bottleneck feature) as the global image descriptor. Also, since HyperPatch-Match is an iterative algorithm, we allowed it to run for 1024 iterations.

\begin{figure*}[t]
\centering
\includegraphics[width=\textwidth]{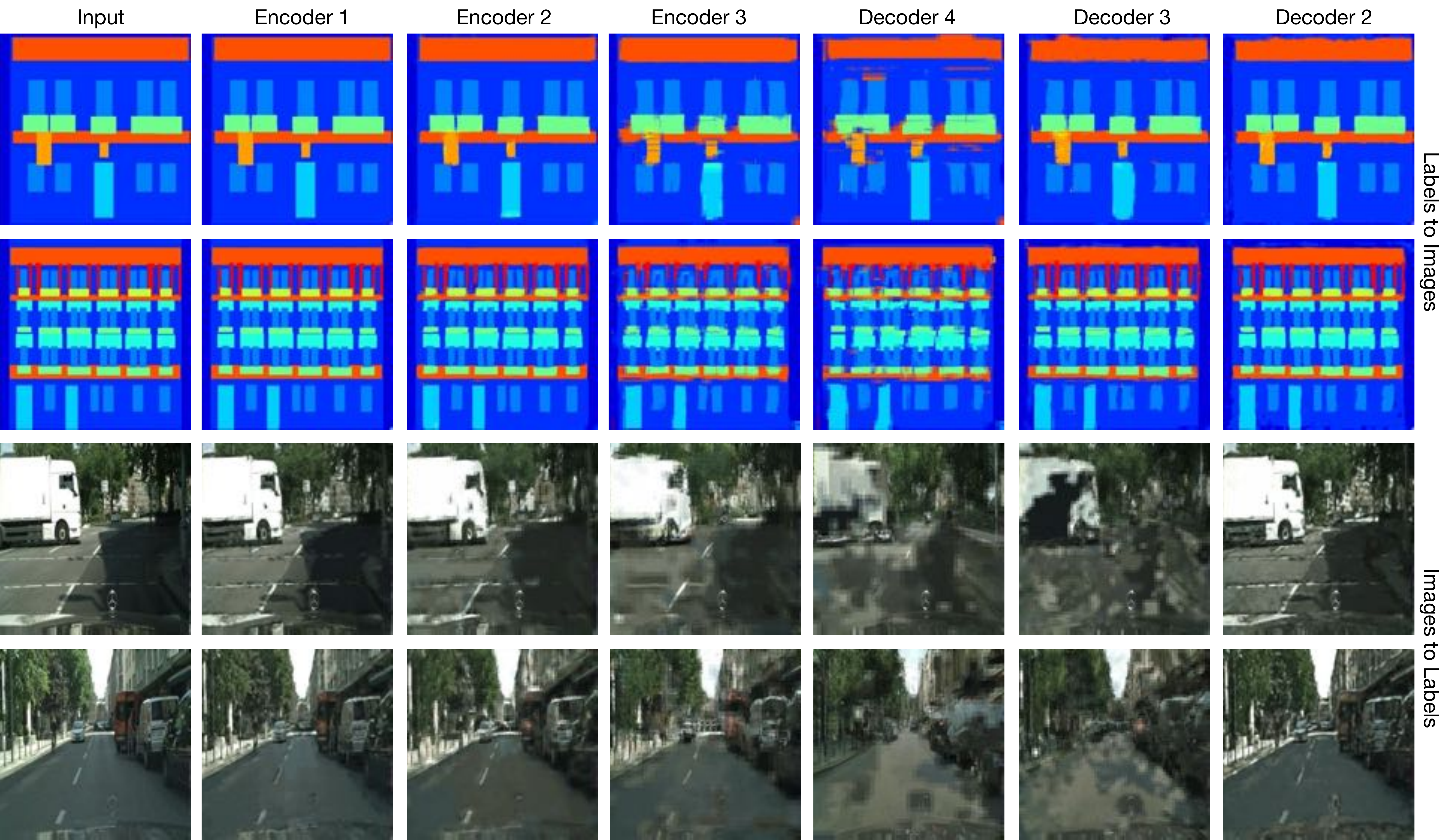}
\vspace{-4mm}
\caption{Input reconstructions using patch representations from encoder and decoder layers. We can observe that the first layers of the encoder and the last layer of the decoder produce a good reconstruction of the input. On the other hand, the reconstructions from the inner layers tend to maintain the structure of the input image, but the quality of the reconstructions decays as we get closer to the middle layers.}
\label{fig:input_reconstructions}
\vspace{-4mm}
\end{figure*}

The results of the input reconstructions are shown in Fig.~\ref{fig:input_reconstructions}. The first two rows show reconstructions for a labels-to-image task, while the bottom two rows show reconstructions for an image-to-labels task. The left column in this Figure shows the inputs to the network, and the remaining columns show reconstructions computed from three layers belonging to the encoder and decoder parts of the CNN. Given that Pix2Pix uses skips connections, i.e., the output of an encoder layer is concatenated to the output of a decoder layer to form the input of the subsequent decoder layer, Fig.~\ref{fig:input_reconstructions} shows reconstructions of the corresponding encoder and decoder layers. This means that the activations of Encoder 1 are part of the activations of Decoder 2.

We can observe in Fig.~\ref{fig:input_reconstructions} that the first layers of the encoder produce a good quality reconstruction of the input, an explanation for this observation is that much of the information to reconstruct the input is still present in the first encoder layer because only a layer of 2D convolutions has been applied. Surprisingly, the last layers of the decoder still produce good quality reconstructions. This can be attributed to the skip connections making both layers (Encoder 1 and Decoder 2) possess the same amount of information to reconstruct a good quality image. On the other hand, the reconstructions from the inner layers tend to maintain the structure of the input image, but the quality  of the reconstructions decays as we get closer to the middle of the U-Net architecture. We refer the reader to the supplemental material for additional input reconstructions. These results show that CompNN is a capable inversion-by-patch-correspondences method. 

\begin{figure*}[t]
\centering
\includegraphics[width=\textwidth]{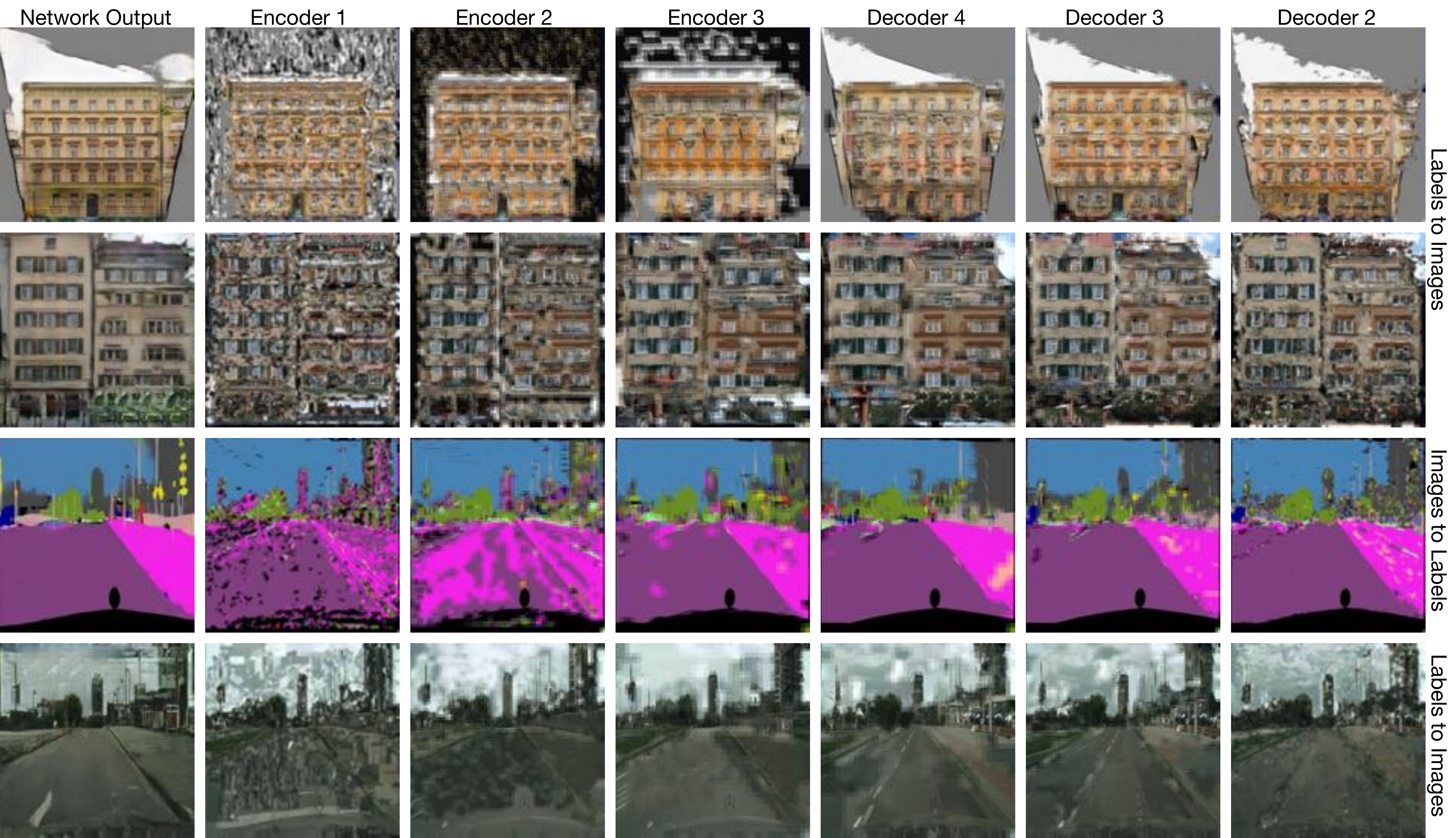}
\vspace{-4mm}
\caption{Output reconstructions using patch representations from encoder and decoder layers. We can observe that the reconstructions from the encoder layers present an overall structure of the output image but lack details that the output image possess. On the other hand, the reconstructions of the decoder layer possess structural information and more details present in the output of the network.}
\label{fig:output_reconstructions}
\vspace{-4mm}
\end{figure*}

The results of the output reconstructions are shown in Fig.~\ref{fig:output_reconstructions}. The column in the left shows the output image generated by the underlying CNN. The organization of the remaining columns is the same as that of the Fig.~\ref{fig:input_reconstructions}. We can observe that the reconstructions from the encoder layers present an overall structure of the output image but lack details that the output image possess. For instance, consider the first row. The reconstructions from the encoder layers preserve the structure of the facade, i.e., the color and location of windows and doors. On the other hand, the reconstructions of the decoder layer possess structural information and more details present in the output of the network. For instance, in the first row, the output image contains a wedge depicting the sky. That part is present in the reconstructions of the decoder layers, but it is not present in the reconstructions of the encoder layers. Despite the fact that we use an approximation method to speed up the nearest neighbor search, we can observe that the reconstruction from the last decoder layer resembles well the image generated by the CNN. Note that sky details appear as well for the second row only in the decoder layers. This suggest that the hallucinations only emerge from the decoder layers of the U-net architecture. More results supporting this hypothesis can be seen in Fig.~\ref{fig:facades_output_recons} in the appendix.

\begin{figure*}[tbp]
\centering
\includegraphics[width=1\linewidth]{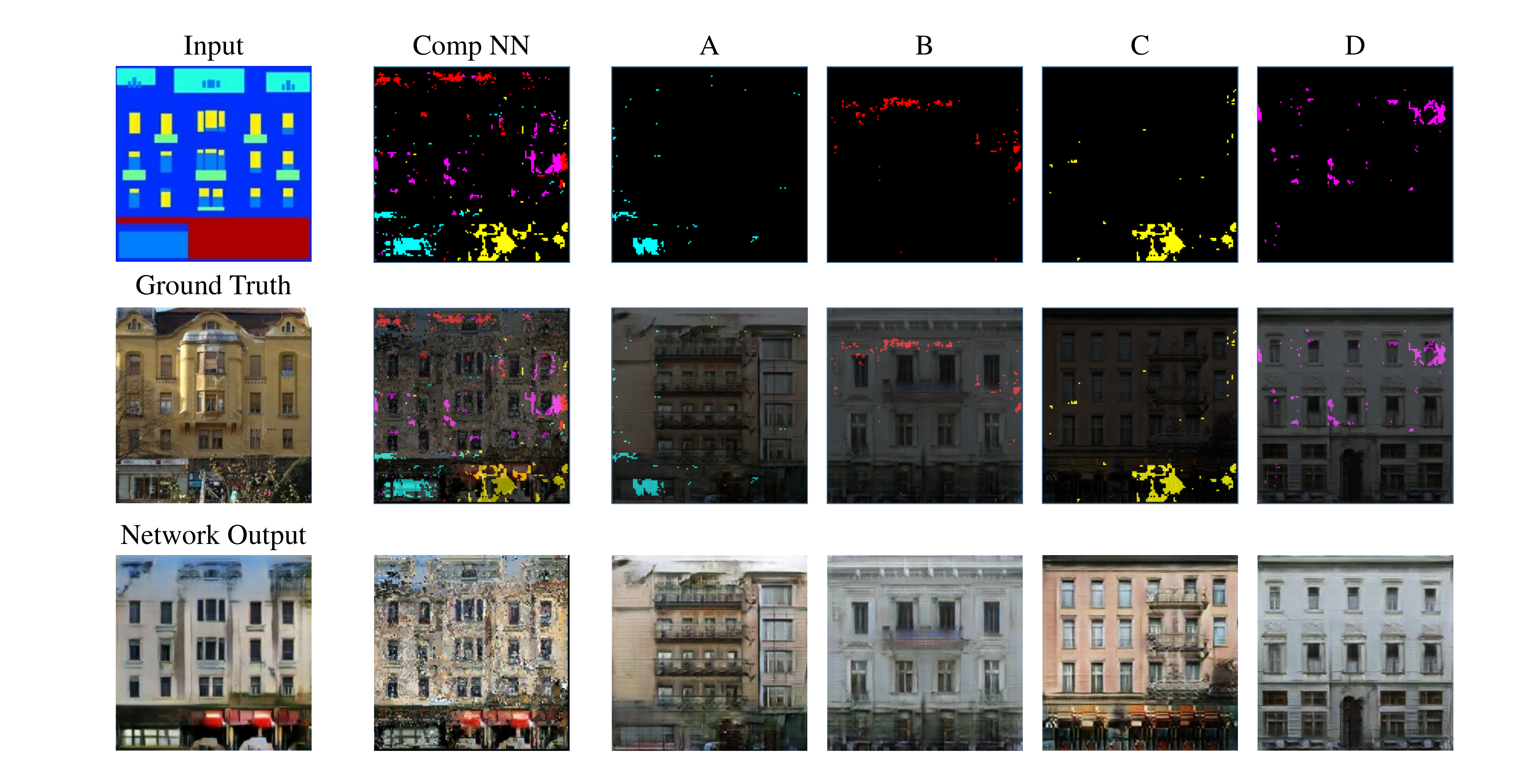}
\vspace{-8mm}
\caption{\textbf{Correspondence map:} Given the input label mask on the top left, we show the ground-truth image and the output of Pix2Pix below. Why does Pix2Pix synthesize the peculiar red awning from the input mask? To provide an explanation, we use CompNN to synthesize an image by explicitly cutting-and-pasting (composing) patches from training images. We color code pixels in the training images to denote correspondences. For example, CompNN copies doors from training image A (blue) and the red awning from training image C (yellow).}
\label{fig:corres_map}
\vspace{-4mm}
\end{figure*}

To interpret the synthesized outputs of CNNs for pixel-level tasks, we show a correspondence map in Fig.~\ref{fig:corres_map} illustrating that CompNN is a good tool for interpreting the outputs of the network. This Figure shows an example of labels-to-image task. The organization of the Figure is the following, in the left column we show the input, ground truth or target image, and the output of the network. In the second column, we color coded the correspondences found that contribute to the synthesis of the output image (shown in the last row in that column). The remaining columns show the different source images that contribute to the composition of the output image. These results suggest that the network learns an embedding that enables a rapid search of source patches that can be used to synthesize a final output. In other words, the embedding encodes patches that can be composed together to synthesize an output image.

Unlike existing interpretation methods that aim to reconstruct the input image only, CompNN is also capable of synthesizing an output image that resembles the generated image by the CNN. This is an important feature because it allows us to interpret the embeddings in charge of generating the synthesized image. Moreover, these results show that the smoothness variation of natural scenes is still present in the activation tensors of both encoder and decoder layers. Thanks to this property, CompNN is able to compute correspondences efficiently using HyperPatch-Match and compose an output image.

To evaluate the reconstructions in a quantitative way, we use the reconstructions from images to labels and assess them by Mean Pixel Accuracy and Mean Intersection over Union (IoU). For this experiment, we consider Pix2Pix on Facades and Cityscapes datasets as well as SegNet~\cite{BadrinarayananK15} on the Camvid dataset for the images-to-labels task. Also, we used HyperPatch-Match to compute correspondences for the Pix2Pix representations, and used an exhaustive search to compute the correspondences for the SegNet representations. The synthesized labeled-images used patch representations from the Decoder 2 layer for the Pix2Pix network, and Decoder 4 or the last layer before the softmax layer for the SegNet network. We trained a SegNet model for the Camvid dataset using a publicly available tensorflow implementation~\footnote{SegNet Implementation: \url{https://github.com/tkuanlun350/Tensorflow-SegNet}}.

The results for the output reconstructions are shown in Table~\ref{tab:output_reconstruction}. The rows of the Table show from top to bottom the metrics for the CNN synthesized image (the baseline); GR (Bottleneck), a reconstruction which simply returns the most similar image in the training set using as global feature the bottleneck activation tensor; GR (FC7), a global reconstructions using the whole activation tensor of the penultimate layer; and output reconstructions (OR) using top 1, 16, 32, and 64 global images as the constrained training set for CompNN and HyperPatch-Match. The result for Camvid dataset which uses exhaustive search is placed in the OR (top 1) row. The entries in the Table show the metric differences between the reconstructions and the baseline, and we show in bold the numbers that are closest to the baseline. We can observe in Table~\ref{tab:output_reconstruction} that the compositional reconstructions overall tend to be close enough to those of the baseline. This is because the absolute value of the differences is small overall. Moreover, we can observe that considering more images in the training set for HyperPatch-Match tends to improve results; compare Facades and Cityscapes columns.

\begin{table}[t]
\caption{Quantitative evidence that CompNN can reconstruct very similar output images when compared with those of the network. The Table shows metric differences between the output of the CNN (the baseline) and CompNN reconstructions (OR) and global reconstructions (GR), which simply return the most similar output image from the training set. Bold entries indicate the closest reconstruction to the baseline.}
\vspace{-4mm}
\setlength{\tabcolsep}{3pt}
\def\arraystretch{1.2}
\center

\scriptsize
\begin{tabular}{c| c| c| c| c| c| c}
\hline
\multirow{2}{*}{Approach}  

& \multicolumn{3}{c|}{\em{(Mean Pixel Accuracy)}} & \multicolumn{3}{c}{\em{(Mean IoU)}} \\

\cline{2-7}
&  Facades & Cityscapes &  Camvid &  Facades & Cityscapes &  Camvid\\

\hline
Baseline CNN & 0.5105 & 0.7618 &  0.7900  & 0.2218 & 0.2097 & 0.4443 \\
\hline
\hline
GR (Bottleneck)  & -0.1963 & -0.1488 & -0.2200 & -0.1437 & -0.0735 & -0.1981 \\
\hline
GR (FC7) & -0.1730 & -0.1333 & -0.1263 & -0.1126 & -0.0702 & -0.1358 \\
\hline
OR (top 1)   & \bf{-0.0102} & -0.0545 & \bf{-0.0350} & -0.0253 &  -0.0277 & \bf{-0.0720} \\
\hline
OR (top 16)  & +0.0324 & -0.0218 & -- & \bf{+0.0214} & +0.0014& -- \\
\hline
OR (top 32)  & +0.0336 & -0.0182 & -- & +0.0232  & \bf{+0.0011} & -- \\
\hline
OR (top 64)  & +0.0343 & \bf{-0.0171} & -- & +0.0246 & +0.0020 & -- \\
\hline
\end{tabular}
\quad\quad\quad
\label{tab:output_reconstruction}
\end{table}

\begin{table}[t]
\caption{Mean Pixel Accuracy (MPA) for input reconstructions. We compare the input reconstructions with the original input label images. These results show that there is a good agreement between the input reconstructions and the original input label images.}
\vspace{-3mm}
\setlength{\tabcolsep}{3pt}
\def\arraystretch{1.2}
\center
\scriptsize
\begin{tabular}{c| c| c | c | c }
\hline
Approach &  IR (top 1)  & IR (top 16) & IR (top 32) & IR (top 64)   \\
\hline
Facades  & 0.723 & 0.837 & 0.844 & 0.846\\
\hline 
Cityscapes & 0.816 & 0.894 & 0.898 & 0.901\\
\hline
\end{tabular} \quad\quad\quad
\label{tab:input_reconstruction}
\end{table}

To evaluate the input reconstructions, we utilized a similar approach used to evaluate the output image reconstructions. However, in this case we compared the input reconstructions with the original input label images only for a Pix2Pix network for labels-to-images task on Facades and Cityscapes datasets. We computed their agreement by means of the Mean Pixel Accuracy (MPA) metric, and we show the results in Table~\ref{tab:input_reconstruction}. Note that MPA compares class labels assigned to every pixel. We can observe that the MPA is overall high ($> 0.7$). In particular, we can observe that considering more top $k$ images in the training set for HyperPatch-Match increases the similarity between the reconstruction and the original input image.

\subsection{Semantic Correspondences and Property-Control in the Output Image}
\label{sec:correspondences}
\begin{figure*}[t]
\centering
\includegraphics[width=\textwidth]{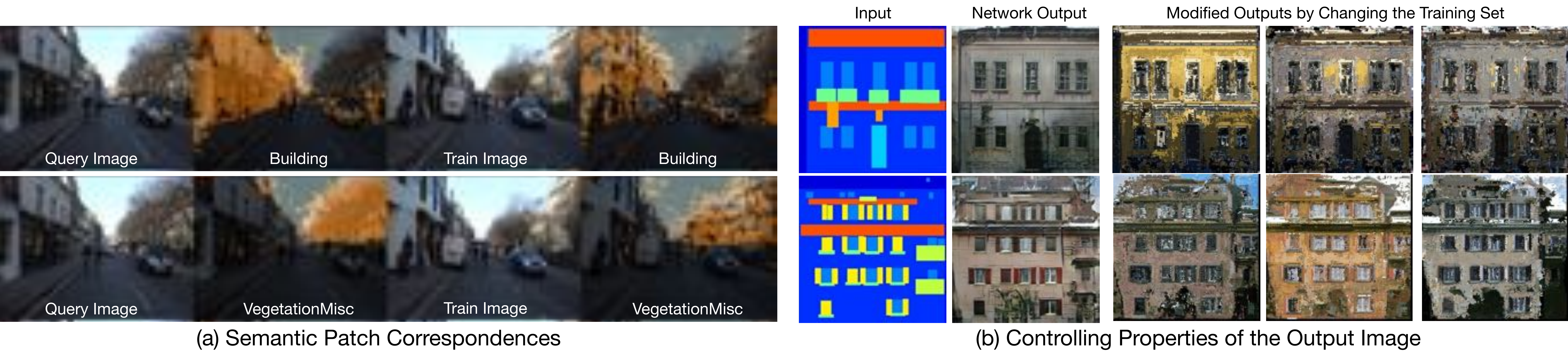}
\vspace{-6mm}
\caption{{\bf (a)} We compute semantic correspondences between pairs of images. To visualize the correspondences, we color-code patches by their semantic label. In general, building patches from one image tend to match building patches from another, and similarly for vegetation. {\bf (b)} We can control the color and window properties of the facades. This feature can be used to understand the bias present in CNNs image outputs and can help users to fix these biases.}
\label{fig:bias}
\vspace{-4mm}
\end{figure*}

Fig.~\ref{fig:bias}(a) shows that HyperPatch-Match can be used to compute semantic correspondences between a pair of images. To show this is possible, we used the patch representations learned by SegNet for the images-to-labels task. To establish the semantic patch correspondences given a pair of images, we first extracted their patch representations from the Decoder 4 layer. Then, we computed the patch correspondences using HyperPatch-Match with 1024 maximum iterations. To visualize the correspondences, we color code patches with their semantic SegNet class. In general, building regions from one image tend to match to building regions from another, and likewise for vegetation.

An additional application of the patch-correspondences is that of controlling properties in the output image. That is, we can control properties (e.g., color of a facade) by simply modifying the images contained in the training set used for computing patch correspondences. To illustrate this, we show results on two different facades output reconstructions in Fig.~\ref{fig:bias}(b). We can observe in both cases that the color of the facades can be manipulated by simply selecting images in the training set depicting facades with the desired color. Also, note that the window frames and sidings changed color and in some cases the type. For instance, in the first row, the third image from left to right, shows windows with white frames and of different style than that of the network output. This control property can help users to understand the bias that CNNs present, and can help users to possibly correct it. See appendix material for additional results.

\section{Conclusions}
\vspace{-2mm}
We have presented compositional nearest neighbors (CompNN), a simple approach based on patch-correspondences to interpret the embeddings learned by an encoder-decoder CNN for spatial predictions or pixel-level tasks. CompNN uses the correspondences that link patches from the input image to image patches in the training set to reconstruct both the CNN's input and output images. Unlike existing interpretation methods that require learning parameters, CompNN is an interpretation-by-example method for encoder-decoder architectures. CompNN generates an image by copying-and-pasting image patches that are composed to generate a coherent one. Thanks to this composition, CompNN is capable of generating images that resemble well both the CNN's input and output images despite its approximate nearest-neighbor algorithm and that the embeddings were trained for deconvolution layers.

We also introduced HyperPatch-Match, an algorithm inspired by Patch-Match~\cite{barnes2009patchmatch} that allows CompNN to efficiently compute patch correspondences. Unlike Patch-Match that uses raw image patches, HyperPatch-Match uses patch-representations that are extracted from the activation tensors of a layer in the underlying CNN. Moreover, unlike Patch-Match and other methods, HyperPatch-Match searches over a database. This is crucial to visualize the embeddings since this database is the training set of the underlying network. We also showed that these patch-representations are useful for establishing semantic correspondences given an image pair, and controlling properties of the reconstructed output image that can help users to understand the present bias in CNNs and possibly correct it.
\begin{appendices}
\section{Introduction}

We present implementation details in Sec.~\ref{sec:impl} and additional results that complement those shown in Section~\ref{sec:experiments}. These include input and output reconstructions in Sec.~\ref{sec:recons}.

\section{Implementation Details}
\label{sec:impl}

The main component requires for the proposed CompNN is to compute patch correspondences. To do this, we implemented a multi-threaded version of an exhaustive search method and the proposed HyperPatch-Match in C++ 11. For linear algebra operations we used Eigen library, and for generating visualizations we used OpenCV 3 via Python wrappers. 

\begin{table}
\caption{Pix2Pix patch and hyperpatch dimensions}
\setlength{\tabcolsep}{12pt}
\centering
\begin{tabular}{c c c}
\hline
Layer & Hyperpatch & Patch \\
\hline
Encoder 1 & $2 \times 2 \times 64$ & $2 \times 2$ \\
Encoder 2 & $2 \times 2 \times 128$ & $4 \times 4$ \\
Encoder 3 & $2 \times 2 \times 256$ & $8 \times 8$ \\
Encoder 4 & $2 \times 2 \times 512$ & $16 \times 16$ \\
Encoder 5 & $2 \times 2 \times 512$ & $32 \times 32$ \\
Encoder 6 & $2 \times 2 \times 512$ & $64 \times 64$ \\
Encoder 7 & $2 \times 2 \times 512$ & $128 \times 128$ \\
Encoder 7 & $2 \times 2 \times 512$ & $128 \times 128$ \\
Decoder 8 & $2 \times 2 \times 1024$ & $128 \times 128$ \\
Decoder 7 & $2 \times 2 \times 1024$ & $64 \times 64$ \\
Decoder 6 & $2 \times 2 \times 1024$ & $32 \times 32$ \\
Decoder 5 & $2 \times 2 \times 1024$ & $16 \times 16$ \\
Decoder 4 & $2 \times 2 \times 512$ & $8 \times 8$ \\
Decoder 3 & $2 \times 2 \times 256$ & $4 \times 4$ \\
Decoder 2 & $2 \times 2 \times 128$ & $2 \times 2$ \\
\hline
\end{tabular}
\label{tab:patch_dims}
\end{table}

We also present the patch sizes and hyperpatch dimensions used for the visualizations of the Pix2Pix embeddings. The parameters are shown in Table~\ref{tab:patch_dims}.

\section{Input and Output Reconstructions}
\label{sec:recons}

In this section, we present additional input and output reconstructions that complement the results shown in Section~\ref{sec:input_output_recon}. 
\subsection{Input Reconstructions}

\begin{figure*}[t]
\centering
\includegraphics[width=\textwidth]{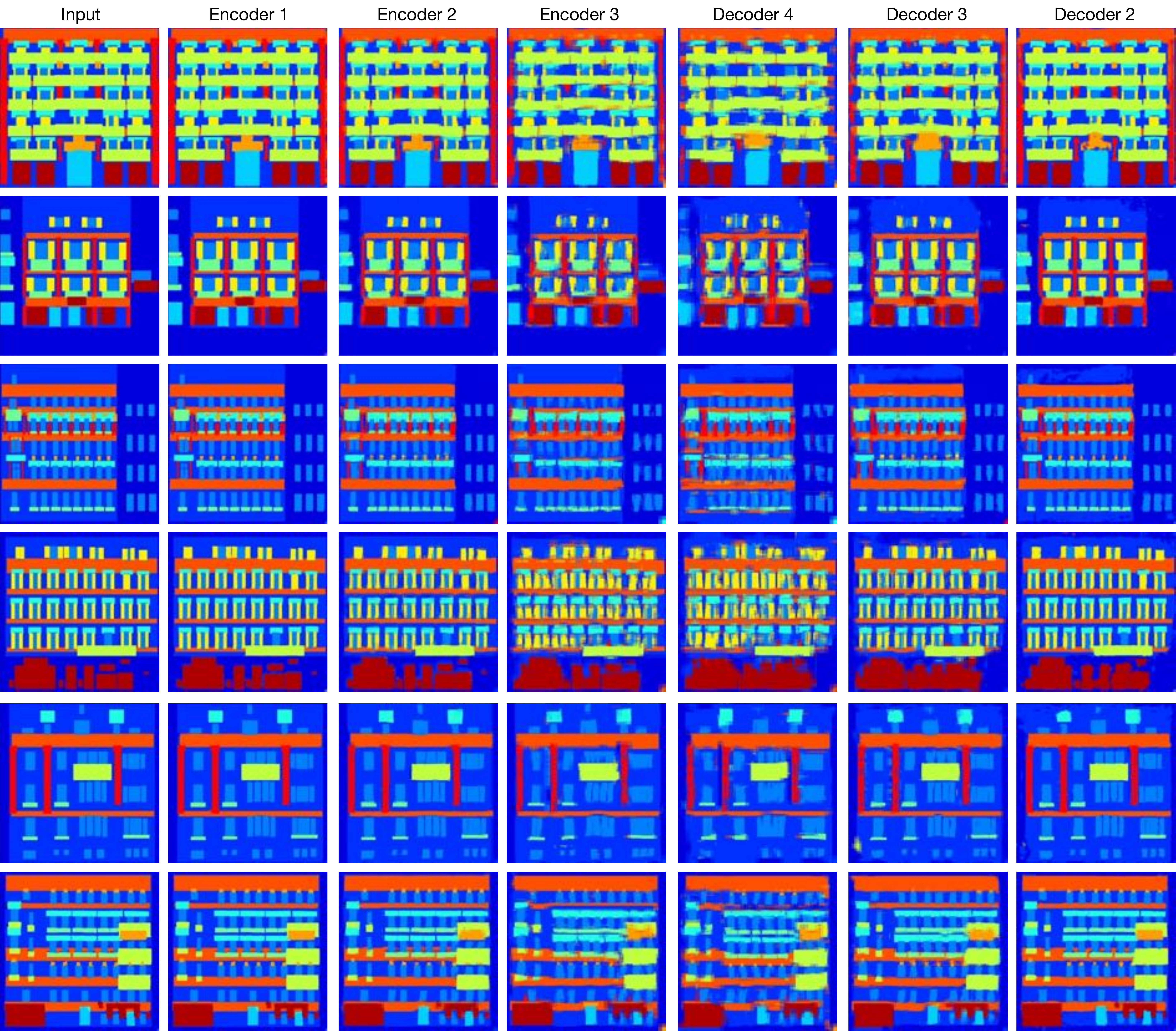}
\caption{Input reconstructions for the labels-to-images task on the Facades dataset. We can observe that the first encoder layers (Encoder 1-2) preserve more information to reconstruct the input image with a good quality. On the other hand, the layers near the bottleneck (e.g., Encoder 3 and Decoder 4-3) lack information to provide a good quality input reconstruction. Note that the last decoder layer (Decoder 2) produces a good input reconstruction despite a few noisy artifacts.}
\label{fig:facades_input_recons}
\end{figure*}

\begin{figure*}[t]
\centering
\includegraphics[width=\textwidth]{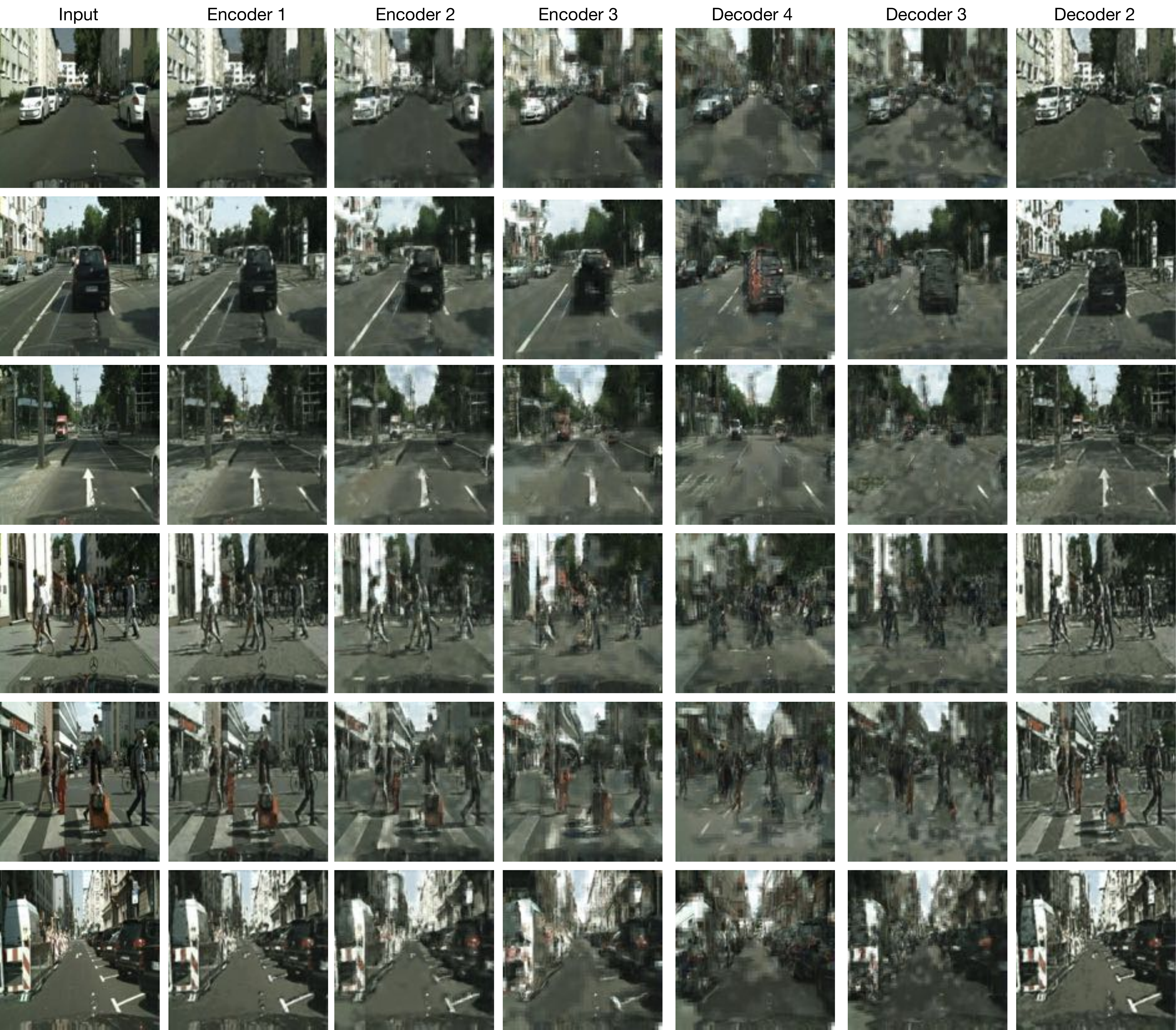}
\caption{Input reconstructions for the images-to-labels task on the Cityscapes dataset. This dataset shows natural scenes with various objects (e.g., humans, cars, buildings). Despite the more complex scenes-to-reconstruct, we observe a similar reconstruction pattern shown in Fig.~\ref{fig:facades_input_recons}. The input reconstructions from Encoder 1 are of higher quality than that of the rest of the layers. The reconstructions decrease quality when they are close to the bottleneck of the network (e.g., Encoder 3, Decoder 4-3). However, we can also observe that the Decoder 2 layer delivers a reasonable reconstruction. We attribute this to the skip connections since they allow a direct transfer of information.}
\label{fig:cityscapes_a2b_input_recon}
\end{figure*}

\begin{figure*}[t]
\centering
\includegraphics[width=\textwidth]{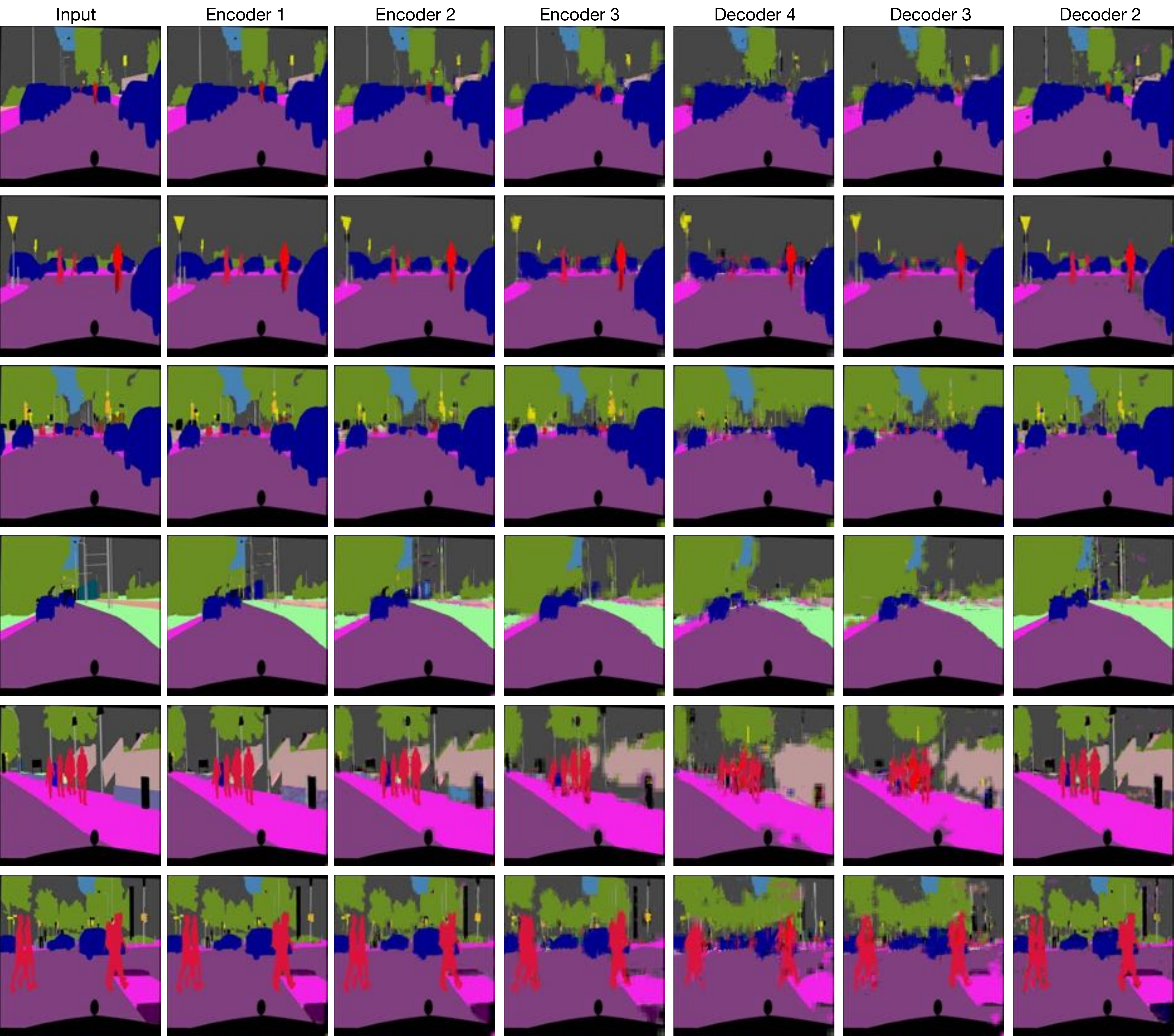}
\caption{Input reconstructions for the labels-to-images task on the Cityscapes dataset. In this dataset we observe the same reconstruction pattern discussed in Fig.~\ref{fig:facades_input_recons} and Fig.~\ref{fig:cityscapes_a2b_input_recon}. We observe that the Encoder 1 layer again produces the highest quality of reconstruction. The Decoder 2 layer produces a comparable reconstruction to those of Encoder 1 thanks to the skip connections. Layers near the bottleneck again show a decrease in quality and we can observe block-like artifacts (see Decoder 4 and 3).}
\label{fig:cityscapes_b2a_input_recon}
\end{figure*}

We show input reconstructions on two validation sets from the Facades and Cityscapes datasets. Similar to the input reconstructions shown in Fig.~\ref{fig:input_reconstructions}, we present reconstructions for the labels-to-images task on the Facades dataset in Fig.~\ref{fig:facades_input_recons}, images-to-labels task on the Cityscapes dataset in Fig.~\ref{fig:cityscapes_a2b_input_recon}, and labels-to-images task on the Cityscapes dataset in Fig.~\ref{fig:cityscapes_b2a_input_recon}. The structure of the Figures is the following: the first column presents the output to reconstruct, while the remaining columns present reconstruction from encoder and decoder layers. Overall, these results show that the first encoder layer (Encoder 1) and the last decoder layer (Decoder 2) produce the highest quality input reconstructions. Also, the reconstructions from layers closer to the bottleneck produce the reconstructions with a decreased quality. These results confirm that the proposed approach is able to reconstruct never-before-seen input images from the patch correspondences and the training set.

\subsection{Output Reconstructions}
\begin{figure*}[t]
\centering
\includegraphics[width=\textwidth]{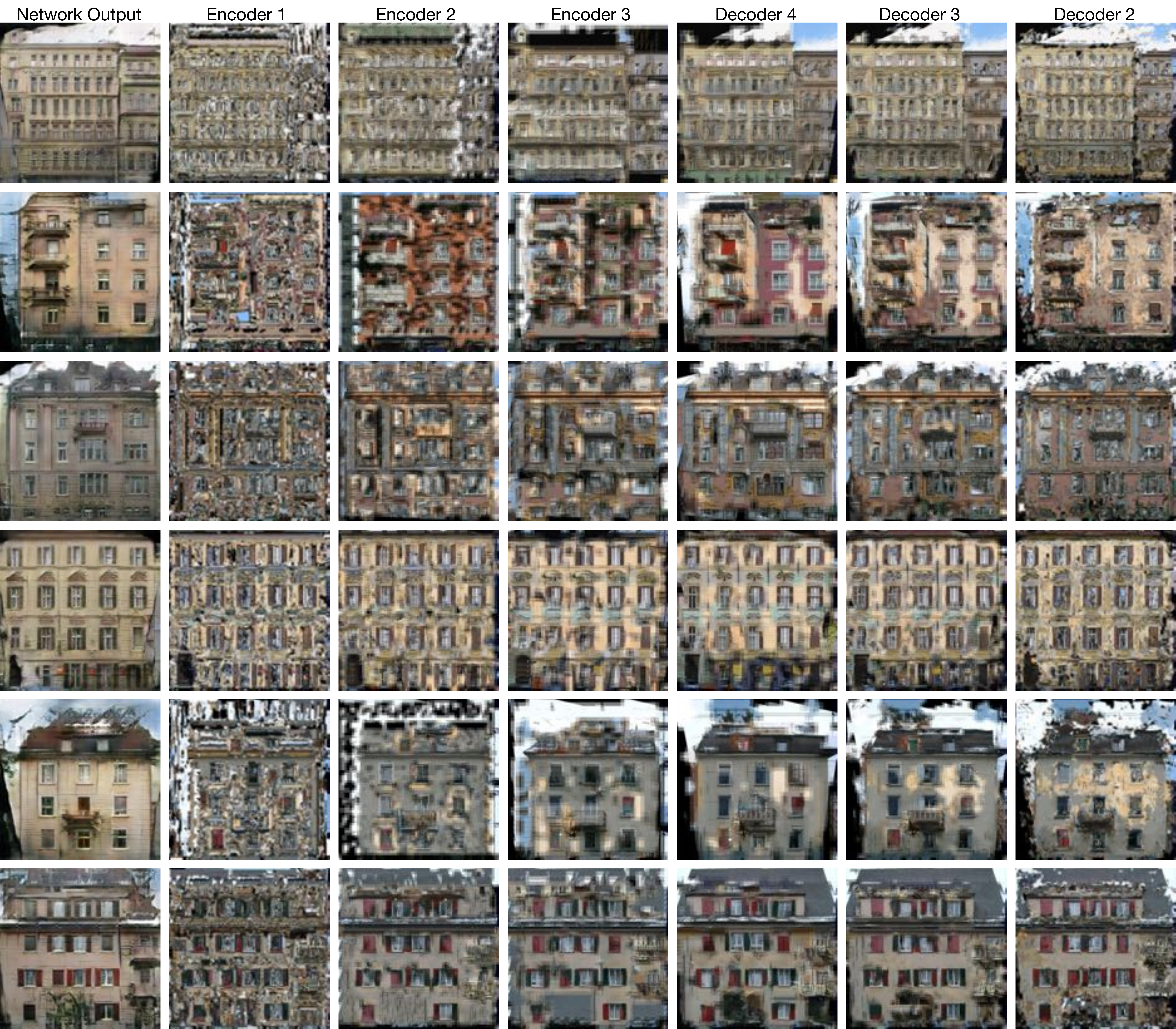}
\caption{Output reconstructions for the labels-to-images task on the Facades dataset. We can observe that the Decoder 2 and Decoder 3 layers produce plausible reconstructions of the output image. The remaining decoder layers struggle more to produce a reasonable reconstruction. Finally, the Encoder layers struggle the most in producing a reasonable reconstruction. We observe that the farther away from the output layer, the higher the struggle to produce a reasonable reconstruction. Note also that hallucination emerge from the Decoder layers. The results of the Decoder 4-2 layers in the first and fifth rows show the inclusion of clouds and blue sky.}
\label{fig:facades_output_recons}
\end{figure*}

\begin{figure*}[t]
\centering
\includegraphics[width=\textwidth]{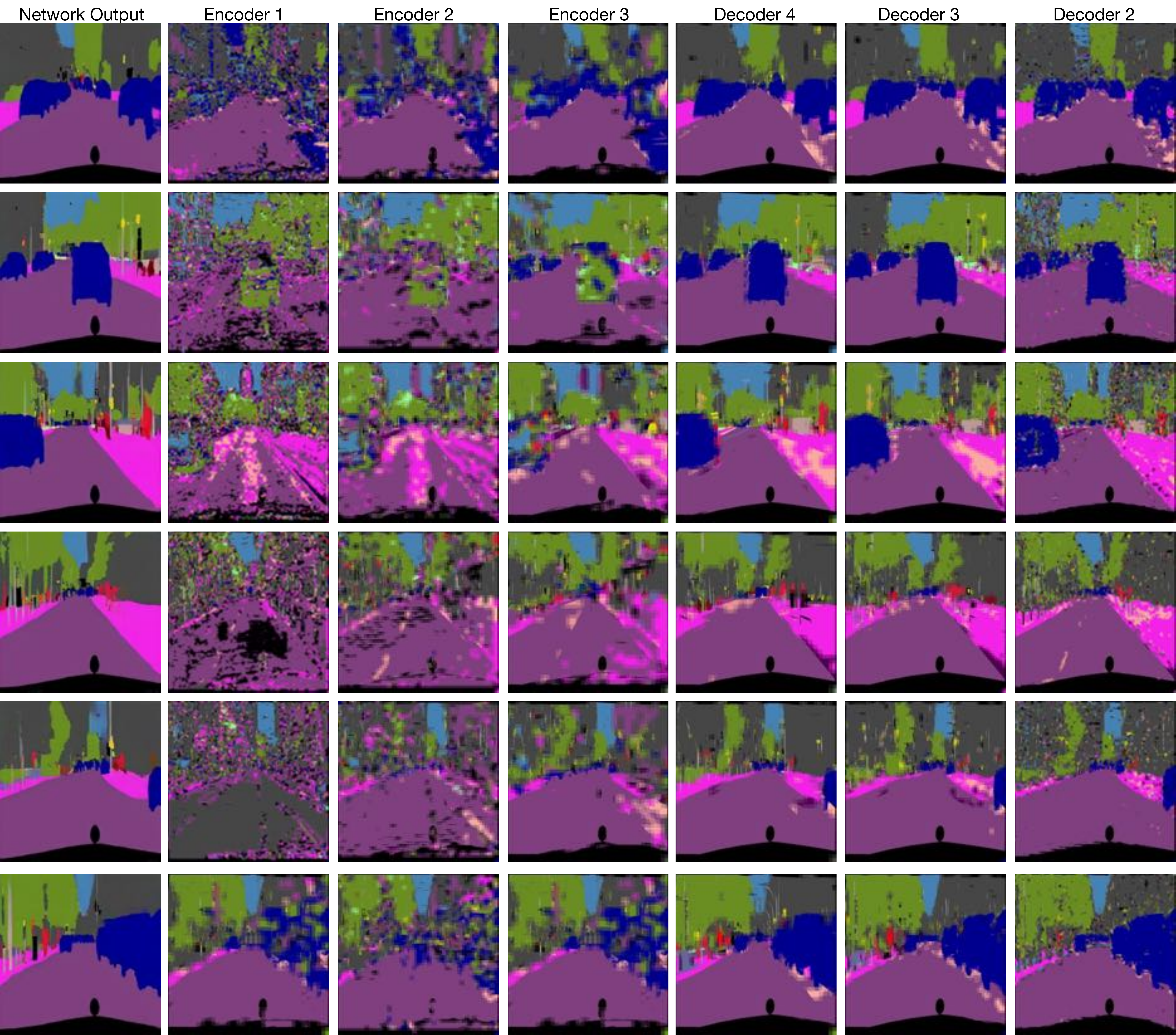}
\caption{Output reconstructions for the images-to-labels task on the Cityscapes dataset. We can observe again that encoder layers struggle the most to reconstruct the output of the network. In contrast with the reconstructions shown in Fig.~\ref{fig:facades_output_recons}, CompNN produced reasonable reconstructions using information from Decoder 4 and 3. In this dataset, the reconstructions from Decoder 2 show noisy artifacts. We attribute this to the approximate nature of CompNN. Despite these artifacts, Decoder-2-based reconstructions resemble well the output of the network.}
\label{fig:cityscapes_a2b_output_recons}
\end{figure*}

\begin{figure*}[t]
\centering
\includegraphics[width=\textwidth]{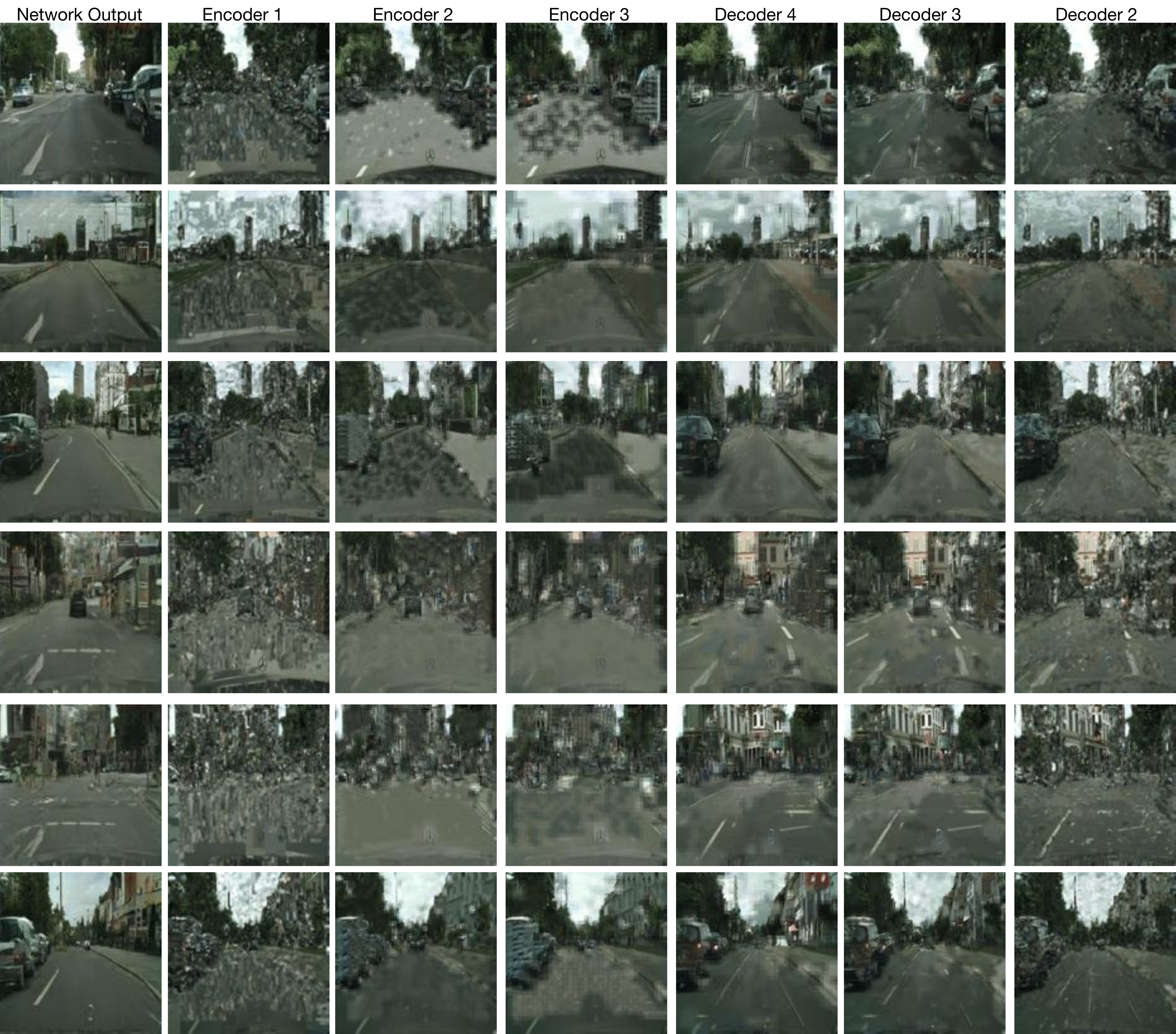}
\caption{Output reconstructions for the labels-to-images task on the Cityscapes dataset. Once again, we can observe that the encoder layers have the least amount of information to reconstruct the structure of the output image well. However, the decoder layers have more information to generate details in the output image that make it closer to the output of the network. Surprisingly, the reconstructions from Decoder 4 and Decoder 3 look cleaner than that of the Decoder 2. The reconstructions of Decoder 2 present noisy artifacts due to the approximate nature of CompNN. Nevertheless, the images resemble well the output of the network.}
\label{fig:cityscapes_b2a_output_recons}
\end{figure*}

We now show output reconstructions on two validation sets from the Facades and Cityscapes datasets. Similar to the output reconstructions shown in Fig.~\ref{fig:output_reconstructions}, we present reconstructions for the labels-to-images task on the Facades dataset in Fig.~\ref{fig:facades_output_recons}, images-to-labels task on the Cityscapes dataset in Fig.~\ref{fig:cityscapes_a2b_output_recons}, and labels-to-images task on the Cityscapes dataset in Fig.~\ref{fig:cityscapes_b2a_output_recons}. The structure of the Figures is the following: the first column presents the output to reconstruct, while the remaining columns present reconstruction from encoder and decoder layers. Different from the input reconstructions, the decoder layers produce better output reconstructions. In particular, the Decoder 2 layer produces the reasonable output reconstructions. On the other hand, the remaining decoder layers maintain the structure of the images but cannot reconstruct the output network image with great detail. Finally, the encoder layers have the least information to generate a plausible reconstruction of the output image of the network. Overall, these results confirm that CompNN is able to generate images that resemble the output of the Network from the patch correspondences and the training set, especially using information from the Decoder 2 layer.

\end{appendices}

{\small
\bibliographystyle{ieee}
\bibliography{compnn}
}

\end{document}